\documentclass{article}

    \usepackage[preprint]{neurips_2019}
    \usepackage{natbib}

\usepackage[utf8]{inputenc} 
\usepackage[T1]{fontenc}    
\usepackage{hyperref}       
\usepackage{url}            
\usepackage{booktabs}       
\usepackage{amsfonts}       

\usepackage{amsmath}
\usepackage{amsthm}
\usepackage{amssymb}
\usepackage{mathtools}

\mathtoolsset{showonlyrefs}  
\usepackage[dvipsnames]{xcolor}
\usepackage[ruled,vlined]{algorithm2e}
\title{Reproducibility Challenge NeurIPS 2019 Report on "Competitive Gradient Descent"}

\author{%
	Gopi Kishan\\
	Computer Science and Engineering\\
	Indian Institute of Technology Roorkee\\
	Roorkee, Uttarakhand \\
	\texttt{gkishan@cs.iitr.ac.in} 
}

\begin{document}
	
	\maketitle
	
	\begin{abstract}
	
	    This is a report for reproducibility challenge of NeurlIPS'19 on the paper \emph{Competitive gradient descent} \citep{Schafer2019CGD}.
	    The paper introduces a novel algorithm for the numerical computation of Nash equilibria of competitive two-player games. It avoids oscillatory and divergent behaviors seen in alternating gradient descent.
	    
	    The purpose of this report is to critically examine the reproducibility of the work by \citep{Schafer2019CGD}, within the framework of the NeurIPS 2019 Reproducibility Challenge. The experiments replicated in this report confirms the results of the original study. Moreover, this project offers a Python(Pytorch based) implementation of the proposed CGD algorithm which can be found at the following public git repository: (\url{https://github.com/GopiKishan14/Reproducibility_Challenge_NeurIPS_2019})

	\end{abstract}

	
	\newcommand{\Reals}{\mathbb{R}}
	\newcommand{\argmin}{\operatorname{argmin}}
	\newcommand{\argmax}{\operatorname{argmax}}
	\newcommand{\Id}{\operatorname{Id}}
	\newcommand{\defeq}{\coloneqq}
	\newcommand{\steps}{\lambda}
	\newcommand{\df}[1]{\operatorname{d} \! #1}
	\newcommand{\todo}[1]{{\color{red} #1}}
	\newcommand{\Expect}{\mathbb{E}}
	\newcommand{\Cov}{\operatorname{Cov}}
	\newcommand{\Var}{\operatorname{Var}}
	\newcommand{\tN}{\tilde{N}}
	\newcommand{\tM}{\tilde{M}}
	\newcommand{\bM}{\bar{M}}
	
	\newtheorem{theorem}{Theorem}[section]
	\newtheorem{corollary}{Corollary}[theorem]
	\newtheorem{lemma}[theorem]{Lemma}
	\newtheorem{remark}[theorem]{Remark}

	\section{Introduction and Motivation}
	The original paper introduces a new algorithm for the numerical computation of Nash equilibria of competitive two-player games.
    Authors suggest their method is a natural generalization of gradient descent to the two-player scenario where the update is given by the Nash equilibrium of a regularized bilinear local approximation of the underlying game. It  avoids oscillatory and divergent behaviors seen in alternating gradient descent. 
    
    \emph{The paper proposes several experiments to establish the robustness of their method. This project aims at replicating their results.}
    
    The paper provides a detailed comparison to methods based on \emph{optimism} and \emph{consensus} on the properties of convergence and stability of various discussed methods using numerical experiments and rigorous analysis.

    
    In order to understand these terms, comparison and proposed method and examine the results of the experiments, next section gives a necessary background of the original paper.

	\section{Background}
	
	The traditional optimization is concerned with a single agent trying to optimize a cost function. It can be seen as $\min_{x \in \Reals^m} f(x)$. The agent has a clear objective to find (“Good local”) minimum of $f$. Gradeint Descent (and its varients) are reliable Algorithmic Baseline for this purpose. 
	
	The paper talks about \textbf{Competitive optimization}.
	Competitive optimization extends this problem to the setting of multiple agents each trying to minimize their own cost function, which in general depends on the actions of all agents.
	
    The paper deals with the case of two such agents:
    \begin{align}
        \label{eqn:game}
        &\min_{x \in \Reals^m} f(x,y),\ \ \ \min_{y \in \Reals^n} g(x,y)
    \end{align}
    for two functions $f,g:\Reals^m \times \Reals^n \longrightarrow \Reals$. \\

In single agent optimization, the solution of the problem consists of the minimizer of the cost function.
In competitive optimization, the right definition of \emph{solution} is less obvious, but often one is interested in computing
Nash-- or strategic equilibria:  Pairs of strategies, such that no player can decrease their costs by unilaterally changing their strategies.
If $f$ and $g$ are not convex, finding a global Nash equilibrium is typically impossible and instead we hope to find a "good" local Nash equilibrium.

\textbf{Gradient descent/ascent and the cycling problem:}
For differentiable objective functions, the most naive approach to solving~\eqref{eqn:game} is gradient descent ascent (GDA), whereby both players independently change their strategy in the direction of steepest descent of their cost function.
Unfortunately, this procedure features oscillatory or divergent behavior even in the simple case of a bilinear game ($f(x,y) = x^{\top} y = -g(x,y)$) (see Figure~\ref{fig:bilinear_strong}).
In game-theoretic terms, GDA lets both players choose their new strategy optimally with respect to the last move of the other player.
Thus, the cycling behaviour of GDA is not surprising: It is the analogue of \emph{"Rock! Paper! Scissors! Rock! Paper! Scissors! Rock! Paper!..."} in the eponymous hand game.
While gradient descent is a reliable basic \emph{workhorse} for single-agent optimization, GDA can not play the same role for competitive optimization. 
At the moment, the lack of such a \emph{workhorse} greatly hinders the broader adoption of methods based on competition.

\subsection{Competitive gradient descent}
\label{sec:cgd}
Authors propose a novel algorithm, which they call \emph{competitive gradient descent} (CGD), for the solution of competitive optimization problems  $\min_{x \in \Reals^m} f(x,y),\ \min_{y \in \Reals^n} g(x,y)$, where they have access to function evaluations, gradients, and Hessian-vector products of the objective functions. \footnote{Here and in the following, all derivatives are evaluated in the point $(x_k, y_k)$}
\begin{algorithm}[H]
	\For{$0 \leq k \leq N-1$}{
              $x_{k+1}  = x_{k} - \eta \left( \Id - \eta^2 D_{xy}^2f D_{yx}^2 g \right)^{-1} 
              \left( \nabla_{x} f - \eta D_{xy}^2f  \nabla_{y} g \right)$\;
              $y_{k+1} = y_{k} - \eta \left( \Id - \eta^2 D_{yx}^2g D_{xy}^2 f \right)^{-1}  
              \left( \nabla_{y} g - \eta D_{yx}^2g  \nabla_{x} f \right)$\;
	}
	\caption{\label{alg:CGD} Competitive Gradient Descent (CGD)}
	\Return $(x_{N},y_{N})$\;
\end{algorithm}
To motivate this algorithm, authors remind us that gradient descent with stepsize $\eta$ applied to the function $f:\Reals^m \longrightarrow \Reals$ can be written as
\begin{equation}
    x_{k+1} = \argmin \limits_{x \in \Reals^m} (x^{\top} - x_{k}^{\top}) \nabla_x f(x_k) + \frac{1}{2\eta} \|x - x_{k}\|^2.
\end{equation}
This models a (single) player solving a local linear approximation of the (minimization) game, subject to a quadratic penalty that expresses her limited confidence in the global accuracy of the model. The natural generalization of this idea to the competitive case should then be given by the two players solving a local approximation of the true game, both subject to a quadratic penalty that expresses their limited confidence in the accuracy of the local approximation. \\
In order to implement this idea, we need to find the appropriate way to generalize the linear approximation in the single agent setting to the competitive setting. 


Authors suggest to use a \emph{bilinear} approximation in the two-player setting.
Since the bilinear approximation is the lowest order approximation that can capture some interaction between the two players, they argue that the natural generalization of gradient descent to competitive optimization is not GDA, but rather the update rule $(x_{k+1},y_{k+1}) = (x_k,y_k) + (x,y)$, where $(x,y)$ is a Nash equilibrium of the game \footnote{We could alternatively use the penalty $(x^{\top}x + y^{\top}y)/(2 \eta)$ for both players, without changing the solution.}
\begin{align}
    \begin{split}
    \label{eqn:localgame}
    \min_{x \in \Reals^m} x^{\top} \nabla_x f &+ x^{\top} D_{xy}^2 f y + y^{\top} \nabla_y f + \frac{1}{2\eta} x^{\top} x \\
    \min_{y \in \Reals^n}  y^{\top} \nabla_y g &+ y^{\top} D_{yx}^2 g x + x^{\top} \nabla_x g + \frac{1}{2\eta} y^{\top} y.
    \end{split}
\end{align}
Indeed, the (unique) Nash equilibrium of the Game~\eqref{eqn:localgame} can be computed in closed form.
\begin{theorem}
    \label{thm:uniqueNash}
    Among all (possibly randomized) strategies with finite first moment, the only Nash equilibrium of the Game~\eqref{eqn:localgame} is given by
    \begin{align}
    \label{eqn:nash}
    &x = -\eta \left( \Id - \eta^2 D_{xy}^2f D_{yx}^2 g \right)^{-1}  
                \left( \nabla_{x} f - \eta D_{xy}^2f  \nabla_{y} g \right) \\
    &y = -\eta \left( \Id - \eta^2 D_{yx}^2g D_{xy}^2 f \right)^{-1}  
                \left( \nabla_{y} g - \eta D_{yx}^2g  \nabla_{x} f \right),
    \end{align}
    given that the matrix inverses in the above expression exist. \footnote{We note that the matrix inverses exist for all but one value of $\eta$, and for all $\eta$ in the case of a zero sum game.}
\end{theorem}

An elegant proof of the above theorem is presented in the original paper \citep{Schafer2019CGD}.

According to Theorem~\ref{thm:uniqueNash}, the Game~\eqref{eqn:localgame} has exactly one optimal pair of strategies, which is deterministic.
Thus, we can use these strategies as an update rule, generalizing the idea of local optimality from the single-- to the multi agent setting and obtaining Algorithm~\ref{alg:CGD}.

\textbf{What I think that they think that I think ... that they do}: Another game-theoretic interpretation of CGD follows from the observation that its update rule can be written as 
\begin{equation}
\label{eqn:whatIthink}
    \begin{pmatrix}
         \Delta x\\
         \Delta y
    \end{pmatrix}
    =  
    -
    \begin{pmatrix}
        \Id        & \eta D_{xy}^2 f \\
        \eta D_{yx}^2 g & \Id        
    \end{pmatrix}^{-1}
    \begin{pmatrix}
         \nabla_{x} f\\
         \nabla_{y} g
    \end{pmatrix}.
\end{equation}
Applying the expansion $ \lambda_{\max} (A) < 1 \Rightarrow \left( \Id - A \right)^{-1} = \lim_{N \rightarrow \infty} \sum_{k=0}^{N} A^k$ to the above equation, we observe that: \\
\begin{itemize}
    \item The first partial sum ($N = 0$) corresponds to the optimal strategy if the other player's strategy stays constant (GDA).
    \item The second partial sum ($N = 1$) corresponds to the optimal strategy if the other player thinks that the other player's strategy stays constant (LCGD, see Figure~\ref{fig:ingredients}).
    \item The third partial sum ($N = 2$) corresponds to the optimal strategy if the other player thinks that the other player thinks that the other player's strategy stays constant, and so forth, until the Nash equilibrium is recovered in the limit.
\end{itemize}

\subsection{Comparison with other methods}
\label{sec:comparison}

As illustrated in Figure~\ref{fig:ingredients}, these six algorithms amount to different subsets of the following four terms.
\begin{figure}
     \begin{align*}
       & \text{GDA: } &\Delta x =  &&&- \nabla_x f&\\
       & \text{LCGD: } &\Delta x =  &&&- \nabla_x f& &-\eta D_{xy}^2 f \nabla_y f&\\
       & \text{SGA: } &\Delta x =  &&&- \nabla_x f& &- \gamma D_{xy}^2 f \nabla_y f&  & & \\
       & \text{ConOpt: } &\Delta x =  &&&- \nabla_x f& &- \gamma D_{xy}^2 f \nabla_y f&  &- \gamma D_{xx}^2 f \nabla_x f& \\
       & \text{OGDA: } &\Delta x \approx &&&- \nabla_x f& &-\eta D_{xy}^2 f \nabla_y f&  &+\eta D_{xx}^2 f \nabla_x f& \\
       & \text{CGD: } &\Delta x = &\left(\Id + \eta^2 D_{xy}^2 f D_{yx}^2 f\right)^{-1}&\bigl( &- \nabla_x f&  &-\eta D_{xy}^2 f \nabla_y f& & & \bigr)
     \end{align*}
    \caption{\label{fig:ingredients} The update rules of the first player for (from top to bottom) GDA, LCGD, ConOpt, OGDA, and CGD, in a zero-sum game ($f = -g$).}
\end{figure}


\begin{enumerate}\label{terms}
    \item \label{item:grad} The \emph{gradient term} $-\nabla_{x}f$, $\nabla_{y}f$ which corresponds to the most immediate way in which the players can improve their cost.
    \item \label{item:comp} The \emph{competitive term} $-D_{xy}f \nabla_yf$, $D_{yx}f \nabla_x f$ which can be interpreted either as anticipating the other player to use the naive (GDA) strategy, or as decreasing the other players influence (by decreasing their gradient).
    \item \label{item:consensus} The \emph{consensus term} $ \pm D_{xx}^2 \nabla_x f$, $\mp D_{yy}^2 \nabla_y f$ that determines whether the players prefer to decrease their gradient ($\pm = +$) or to increase it ($\pm = -$). The former corresponds the players seeking consensus, whereas the latter can be seen as the opposite of consensus. \\
    (It also corresponds to an approximate Newton's method. \footnote{Applying a damped and regularized Newton's method to the optimization problem of Player 1 would amount to choosing $x_{k+1} = x_{k} - \eta(\Id + \eta D_{xx}^2)^{-1} f \nabla_x f \approx x_{k} - \eta( \nabla_xf - \eta D_{xx}^{2}f \nabla_x f)$, for $\|\eta D_{xx}^2f\| \ll 1$.})
    \item \label{item:equilibrium} The \emph{equilibrium term} $(\Id + \eta^2 D_{xy}^2 D_{yx}^2 f)^{-1}$, $(\Id + \eta^2 D_{yx}^2 D_{xy}^2 f)^{-1}$, which arises from the players solving for the Nash equilibrium. 
    This term lets each player prefer strategies that are less vulnerable to the actions of the other player.
\end{enumerate}

\textbf{Further Discussion:}\\
\\
\textbf{\emph{Consensus optimization }(ConOpt)}  \citep{mescheder2017numerics}, penalises the players for non-convergence by adding the squared norm of the gradient at the next location, $\gamma \|\nabla_x f(x_{k+1},y_{k+1}), \nabla_x f(x_{k+1},y_{k+1})\|^2$ to both pla yer's loss function (here $\gamma \geq 0$ is a hyperparameter).\\
\citep{daskalakis2017training} proposed to modify GDA as
\begin{align}
    \label{eqn:updateOGDA}
    \Delta x &= - \left( \nabla_x f(x_{k},y_{k}) 
        + \left( \nabla_x f(x_{k},y_{k}) - \nabla_x f(x_{k-1},y_{k-1}) \right) \right) \\
    \Delta y &= - \left( \nabla_y g(x_{k},y_{k}) 
        + \left( \nabla_y g(x_{k},y_{k}) - \nabla_y g(x_{k-1},y_{k-1}) \right) \right),
\end{align}
which we will refer to as \textbf{\emph{optimistic gradient descent ascent} (OGDA)}.
By interpreting the differences appearing in the update rule as finite difference approximations to Hessian vector products, we see that (to leading order) OGDA corresponds to yet another second order correction of GDA (see Figure~\ref{fig:ingredients}).\\

It will also be instructive to compare the algorithms to \textbf{\emph{linearized competitive gradient descent} (LCGD)}, which is obtained by skipping the matrix inverse in CGD (which corresponds to taking only the leading order term in the limit $\eta D_{xy}^2f \rightarrow 0$).

\section{Experiments and Replications}\label{exp}
This section contains the contribution from this project and defines the experiment.\\
A link to public Github repository (\url{https://github.com/GopiKishan14/Reproducibility_Challenge_NeurIPS_2019}) describes the implementation details of \emph{Competitive Gradient Descent}.


A brief discussion on the implementation of CGD by Authors.\\
\textbf{Computing Hessian vector products:}
First, our algorithm requires products of the mixed Hessian $v \mapsto D_{xy}f v$, $v \mapsto D_{yx}g v$, which we want to compute using automatic differentiation.


Many AD frameworks, like Autograd (\url{https://github.com/HIPS/autograd}) and ForwardDiff(\url{https://github.com/JuliaDiff/ForwardDiff.jl}, \citep{revels2016forward}) together with ReverseDiff(\url{https://github.com/JuliaDiff/ReverseDiff.jl}) support this procedure.

\textbf{Matrix inversion for the equilibrium term}: 
Similar to a \emph{truncated Newton's method} \citep{nocedal2006numerical}, we propose to use iterative methods to approximate the inverse-matrix vector products arising in the equilibrium term~\ref{item:equilibrium}.
We will focus on zero-sum games, where the matrix is always symmetric positive definite, making the conjugate gradient (CG) algorithm the method of choice. 
We suggest terminating the iterative solver after a given relative decrease of the residual is achieved ($\| M x - y \| \leq \epsilon \|x\|$ for a small parameter $\epsilon$, when solving the system $Mx = y$).
In our experiments we choose $\epsilon = 10^{-6}$.
Given the strategy $\Delta x$ of one player, $\Delta y$ is the optimal counter strategy which can be found without solving another system of equations. 

\subsection{Experiment 1}
\label{exp:exp1}
The author suggest that each of the terms (section \ref{terms}) used in the definition of the algorithm is responsible for a different feature of the corresponding algorithm which can illustrated by applying the algorithms to three prototypical test cases.

\begin{figure}
    \centering
    \includegraphics[scale=0.21]{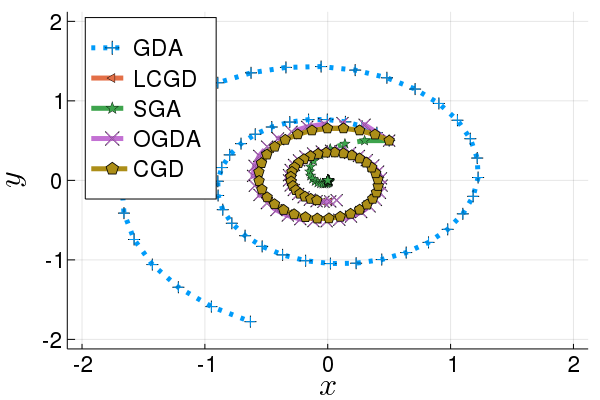}
    \includegraphics[scale=0.21]{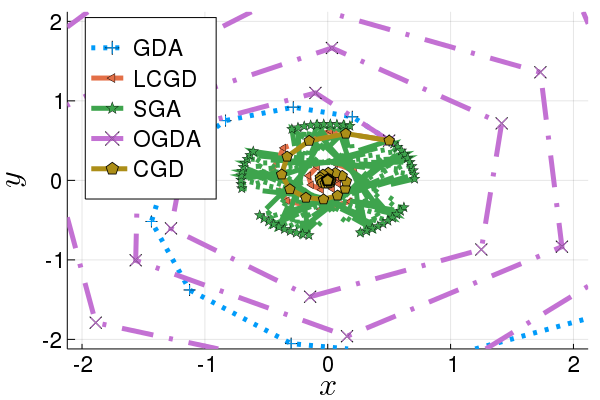}
    \includegraphics[scale=0.21]{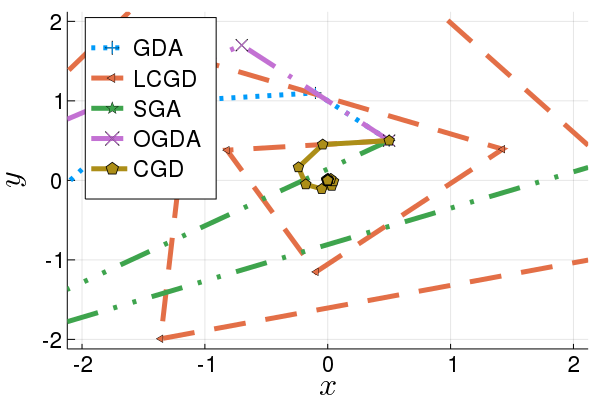}
    \caption{Original Paper Result}
    \includegraphics[scale=0.21]{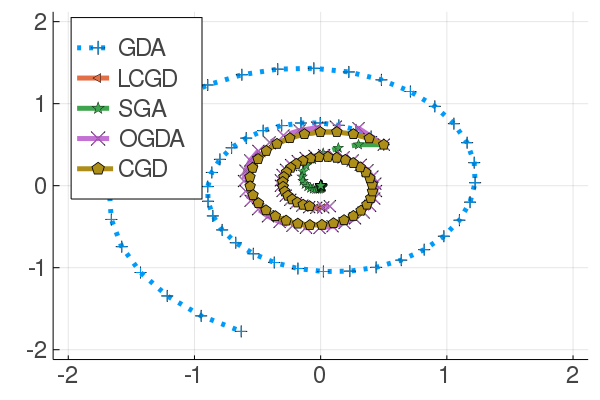}
    \includegraphics[scale=0.21]{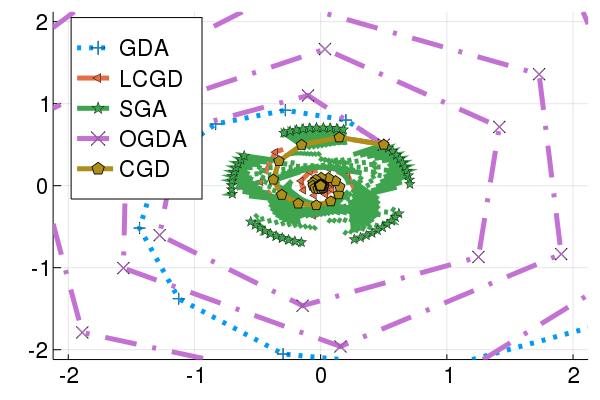}
    \includegraphics[scale=0.21]{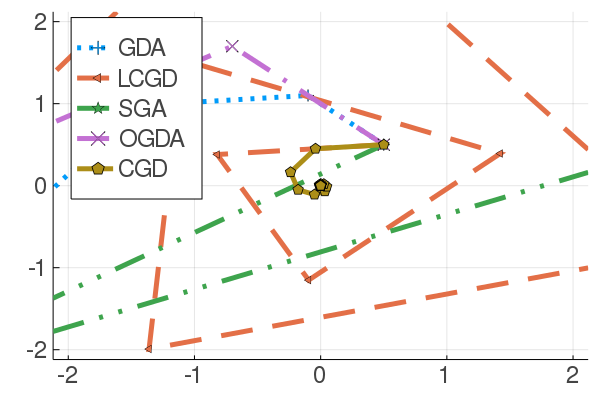}
    \caption{Replicated Result: The first 50 iterations of GDA, LCGD, ConOpt, OGDA, and CGD with parameters $\eta = 0.2$ and $\gamma = 1.0$. 
    The objective function is $f(x,y) = \alpha x^{\top}y$ for, from left to right, $\alpha \in \{1.0, 3.0, 6.0\}$. (Note that ConOpt and SGA coincide on a bilinear problem)}

    \label{fig:bilinear_strong}
\end{figure}

\begin{figure}
    \centering
    \includegraphics[scale=0.21]{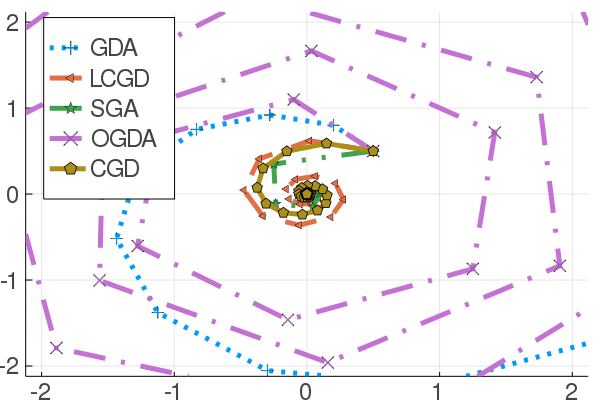}
    \includegraphics[scale=0.21]{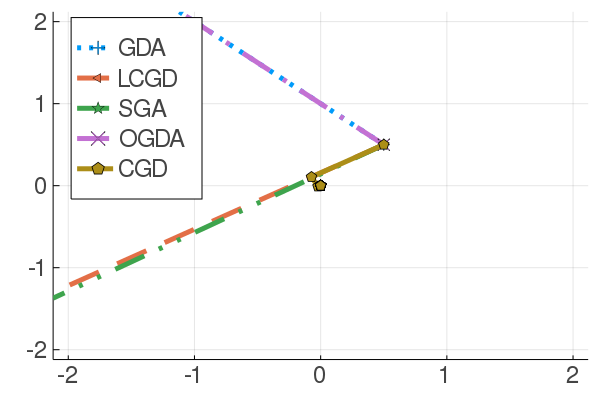}
    \includegraphics[scale=0.21]{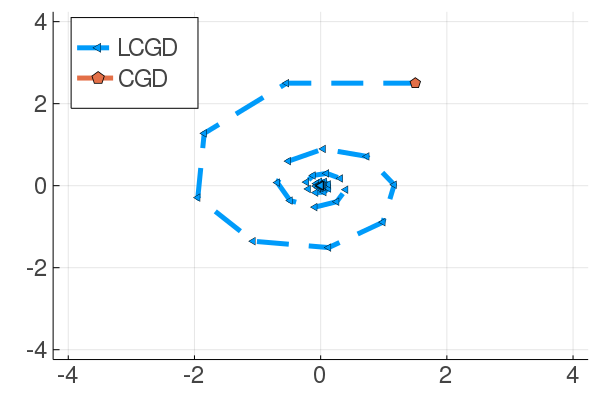}
    \caption{Results for Experiment \ref{exp:exp1} Test \ref{item:test1}. (Refer to the part for explanation)}
    \label{fig:test1}
\end{figure}

\begin{figure}
    \centering
    \includegraphics[scale=0.105]{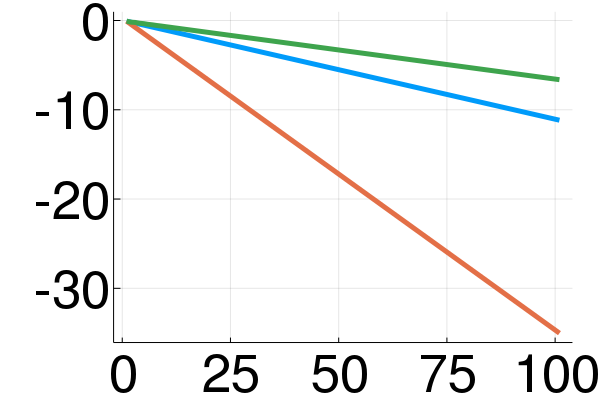}
    \includegraphics[scale=0.105]{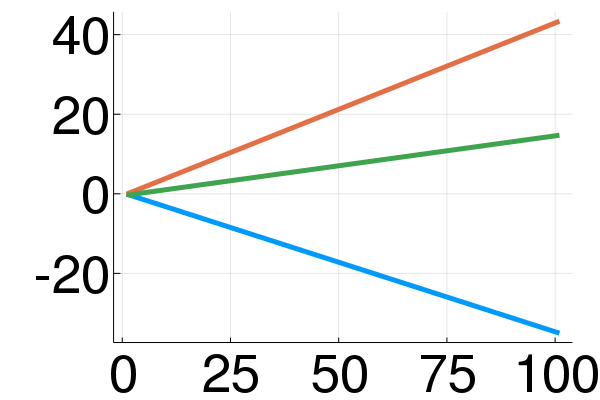}
    \includegraphics[scale=0.105]{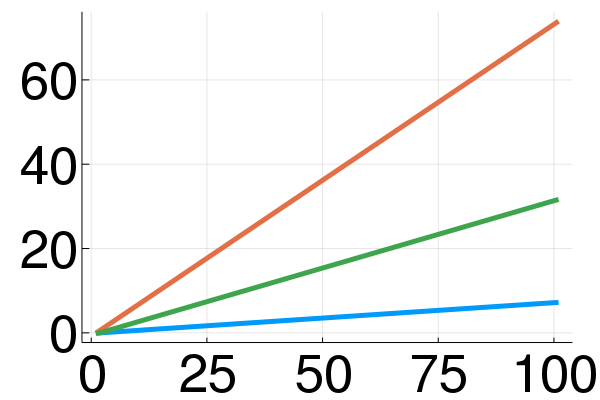}
    \includegraphics[scale=0.105]{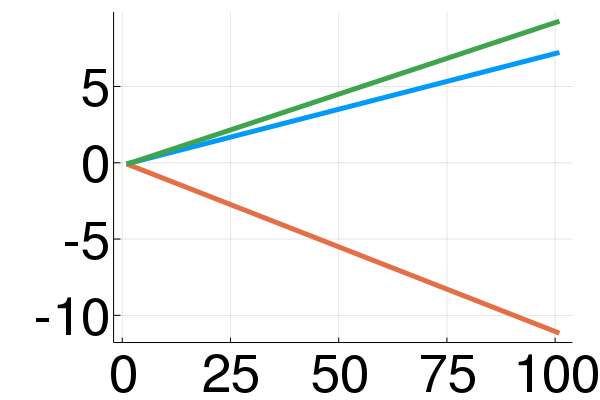}
    \includegraphics[scale=0.105]{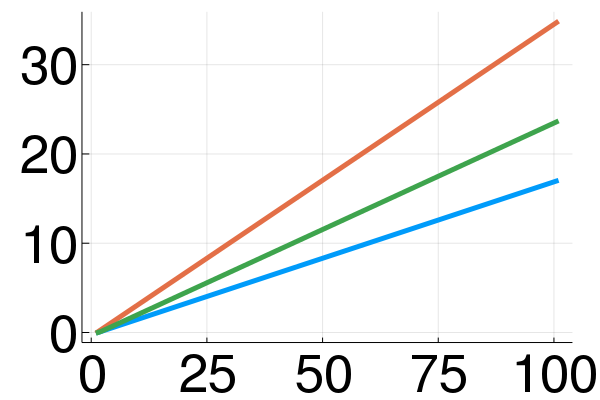}
    \includegraphics[scale=0.105]{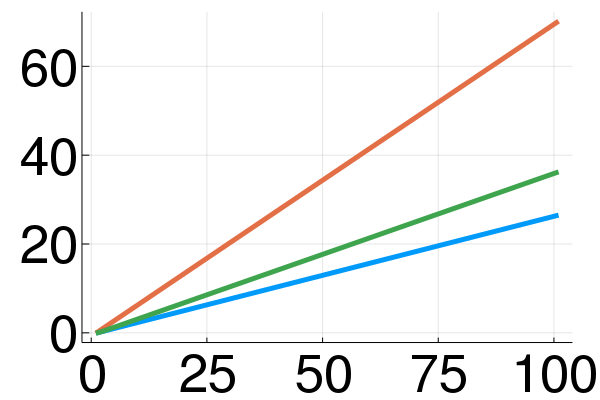}
    \caption{Original Paper result}
    
    \includegraphics[scale=0.105]{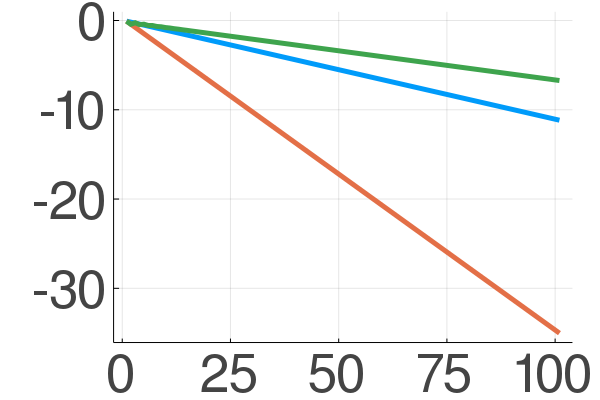}
    \includegraphics[scale=0.105]{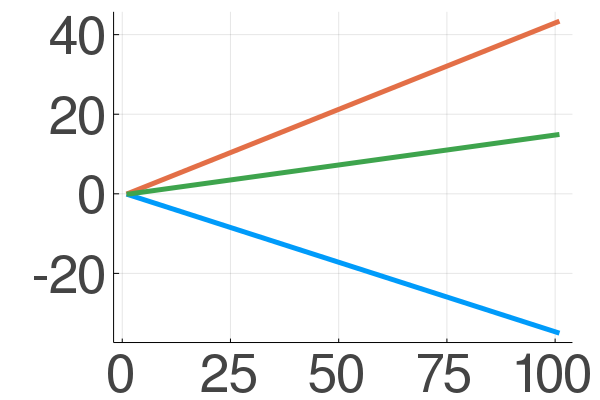}
    \includegraphics[scale=0.105]{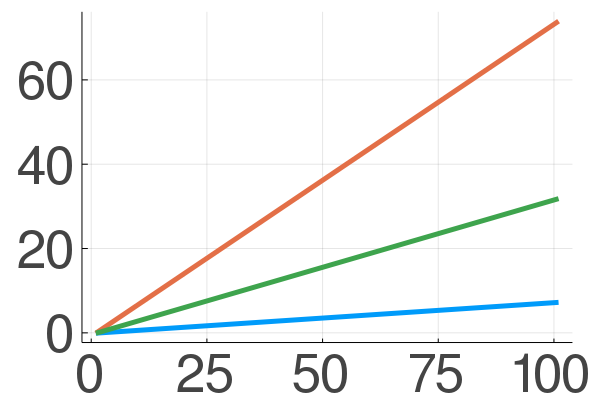}
    \includegraphics[scale=0.105]{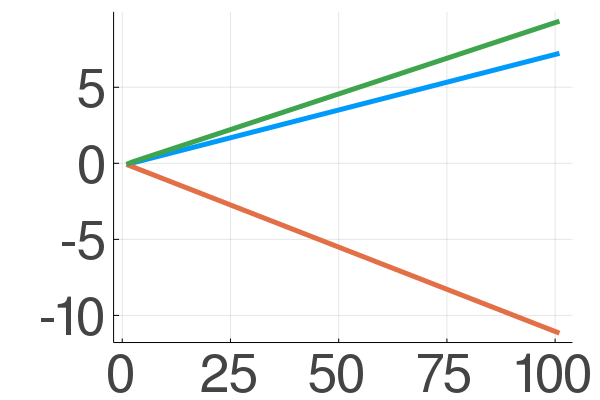}
    \includegraphics[scale=0.105]{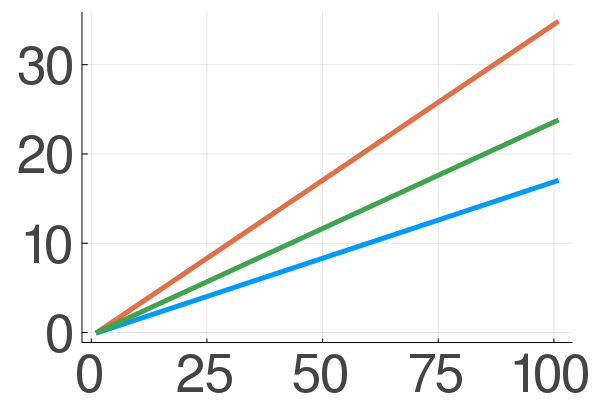}
    \includegraphics[scale=0.105]{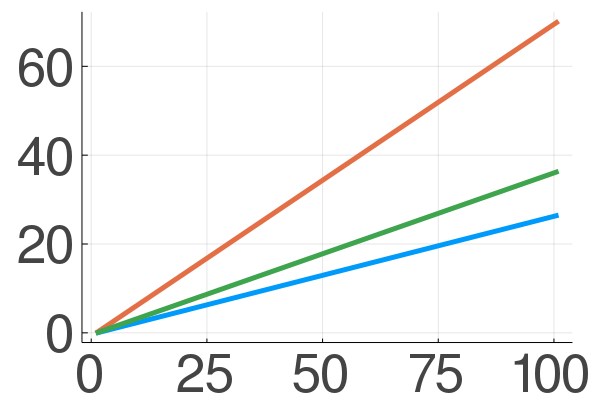}
    
    \caption{Replicated result: 
    We measure the (non-)convergence to equilibrium in the separable convex-concave-- ($f(x,y) = \alpha( x^2 - y^2 )$, left three plots) and concave convex problem ($f(x,y) = \alpha( -x^2 + y^2 )$, right three plots), for $\alpha \in \{1.0,3.0,6.0\}$. 
    (Color coding given by \textcolor{Cyan}{ GDA, SGA, LCGD, CGD}, \textcolor{RedOrange}{ConOpt}, \textcolor{Green}{OGDA}, the y-axis measures  $\log_{10}(\|(x_{k},y_{k})\|)$ and the x-axis the number of iterations $k$.
    Note that convergence is desired for the first problem, while \emph{divergence} is desired for the second problem.}
    \label{fig:quad}
\end{figure}

\begin{figure}
    \centering
    \includegraphics[scale=0.215]{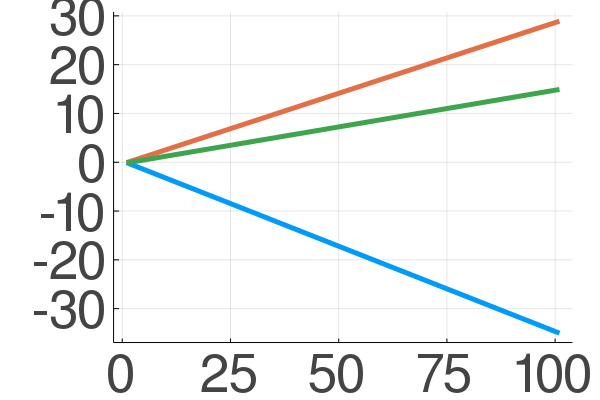}
    \includegraphics[scale=0.215]{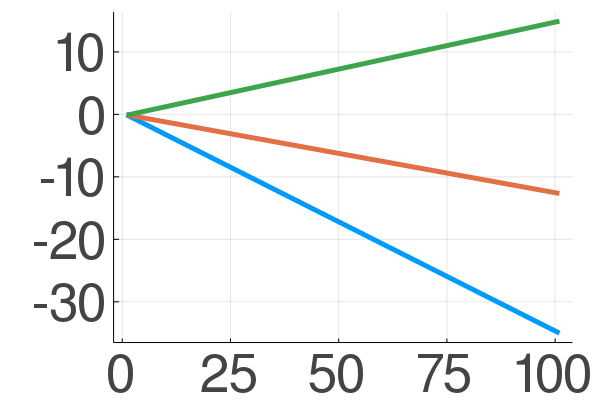}
    \includegraphics[scale=0.215]{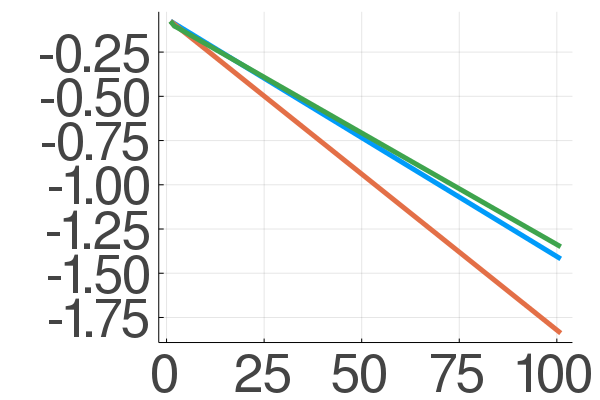}

    \caption{Results for Experiment \ref{exp:exp1} Test \ref{item:test2}. Shows the robustness of CGD, while other methods can also be tuned for convergence at $\alpha = 3.0$ and $\eta = 0.2$ on decreasing $\gamma.$ (convergence desired)}
    \label{fig:test2}

    \includegraphics[scale=0.215]{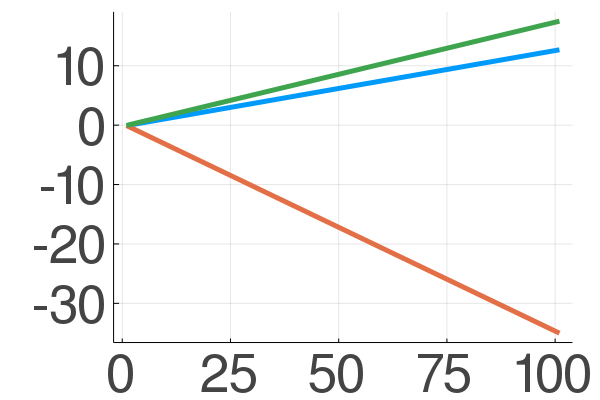}
    \includegraphics[scale=0.215]{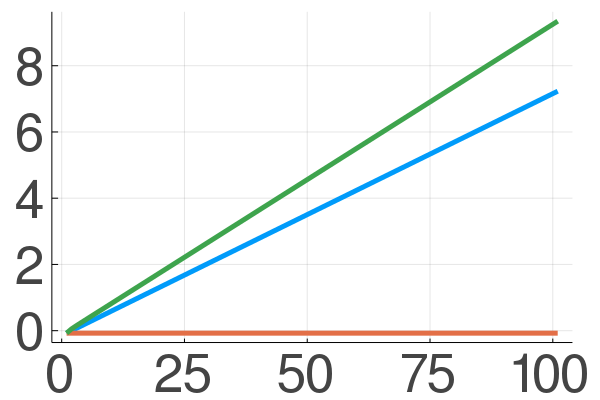}
    \includegraphics[scale=0.215]{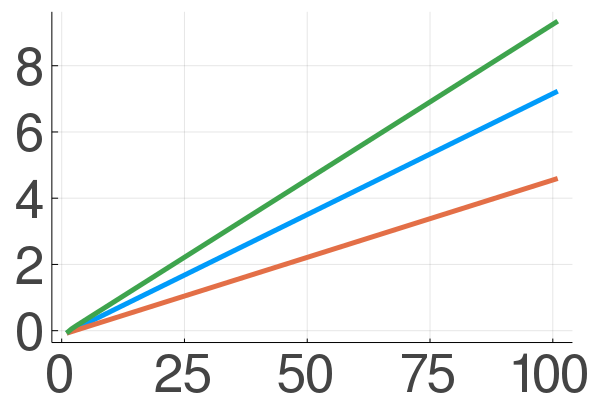}
    
    \caption{Results for Experiment \ref{exp:exp1} Test \ref{item:test3}. Shows the robustness of CGD, while \textcolor{RedOrange}{ConOpt} can also be tuned for convergence at $\alpha = 1.0$ and $\eta = 0.2$ on decreasing $\gamma.$. $\gamma \in \{1.0,0.5,0.2\}$ (divergence desired)}
    \label{fig:test3}
\end{figure}

\begin{enumerate}\label{test}
    \item \label{item:test1}\textbf{Test Case 1:}  We first consider the bilinear problem $f(x,y) = \alpha xy$ (see Figure~\ref{fig:bilinear_strong}).\\
    \textbf{Replication and Findings:}
        As suggested, GDA fails on this problem, for any value of stepsize $\eta$. \\
        For $\alpha = 1.0$, all the other methods converge exponentially towards the equilibrium, with ConOpt and SGA converging at a faster rate.\\
        For $\alpha = 3.0$, OGDA diverges, while ConvOpt and SGA begin to oscillate widely.\\
        For $\alpha = 6.0$, all methods but CGD diverge. This result is compared and summed up in Figure (\ref{fig:bilinear_strong})\\
        However, for $\alpha = 3.0$, if we decrease $\gamma \ to \ 0.5 $ ConOpt and SGA converges. ConOpt and SGA does not converge for any value of $\gamma$ if $\alpha = 6.0$. Figure \ref{fig:test1} Left.\\
        For $\alpha = 6.0$, if stepsize $\eta$ is increased to $\eta = 0.9$, CGD converges faster. Hence, establishing the claim that CGD convergence rate increases with increase in stepsize $\eta$. Figure \ref{fig:test1} Middle.
        
        Interestingly, CGD fails for the problem if initial points set on $y =  5/3 x$ for $\alpha = 3.0$ and $\eta = 0.2$. But LCGD still converges. Figure (\ref{fig:test1}) Right.
        
        


    \item \label{item:test2}\textbf{Test case 2:} In order to explore the effect of the consensus Term~\ref{item:consensus}, we now consider the convex-concave problem $f(x,y) = \alpha(x^2 - y^2)$ (see Figure~\ref{fig:quad}). \\
    \textbf{Replication and Findings:}
    For $\alpha = 1.0$, all algorithms converge at an exponential rate, with ConOpt converging the fastest, and OGDA the slowest.
    
    As we increase $\alpha$ to $\alpha = 3.0$, the OGDA and ConOpt start failing (diverge), while the remaining algorithms still converge at an exponential rate. 
    Upon increasing $\alpha$ further to $\alpha = 6.0$, all algorithms diverge. (See Figure~\ref{fig:quad})\\
    Figure \ref{fig:test2} shows the robustness of CGD, while other methods can also be tuned for convergance at $\alpha = 3.0$ and $\eta = 0.2$ on decreasing $\gamma.$ 
    
    \item \label{item:test3}\textbf{Test case 3:} We further investigate the effect of the consensus Term~\ref{item:consensus} by considering the concave-convex problem $f(x,y) = \alpha( -x^2 + y^2)$ (see Figure~\ref{fig:quad}).\\
    \textbf{Replication and Findings:}
    The critical point $(0,0)$ does not correspond to a Nash-equilibrium, since both players are playing their \emph{worst possible strategy}. 
    Thus it is highly undesirable for an algorithm to converge to this critical point.\\
    However for $\alpha = 1.0$, ConOpt does converge to $(0,0)$ which provides an example of the consensus regularization introducing spurious solutions.
    The other algorithms, instead, diverge away towards infinity, as would be expected.
    In particular, we see that SGA is correcting the problematic behavior of ConOpt, while maintaining its better convergence rate in the first example.
    As we increase $\alpha$ to $\alpha \in \{3.0,6.0\}$, the radius of attraction of $(0,0)$ under ConOpt decreases and thus ConOpt diverges from the starting point $(0.5,0.5)$, as well. (See Figure~\ref{fig:quad})\\
    Figure \ref{fig:test3} shows the robustness of CGD, while \textcolor{RedOrange}{ConOpt} can also be tuned for convergence at $\alpha = 1.0$ and $\eta = 0.2$ on decreasing $\gamma$ $\gamma \in \{1.0,0.5,0.2\}$ (divergence desired).
    
\end{enumerate}

\textbf{Conclusion:}
The first test case illustrates that this is not just a lack of theory, but corresponds to an actual failure mode of the existing algorithms.
Introducing the competitive term (\ref{item:comp}) is enough to fix the cycling behaviour of GDA, OGDA and ConOpt (for small enough $\eta$).

In the Test Case \ref{item:test2} (where convergence is desired), OGDA converges in a smaller parameter range than GDA and SGA, while only diverging slightly faster in the Test Case \ref{item:test3} (where divergence is desired).
ConOpt, on the other hand, converges faster than GDA in the Test Case \ref{item:test2}, for $\alpha = 1.0$ however, it diverges faster for the remaining values of $\alpha$ and, what is more problematic, it converges to a spurious solution in the Test Case \ref{item:test3} for $\alpha = 1.0$.

Based on the findings of Test Case \ref{item:test2} and \ref{item:test3}, the consensus term (\ref{item:consensus}) with either sign does not seem to systematically improve the performance of the algorithm (see Fig \ref{fig:test2} and \ref{fig:test3}), which is why the authors suggest to only use the competitive term (that is, use LOLA/LCGD, or CGD, or SGA).

\subsection{Experiment 2: Estimating a covariance matrix}

To show that CGD is also competitive in terms of computational complexity, authors consider the noiseless case of the covariance estimation example used by \cite{daskalakis2017training}.

Authors consider the problem $-g(V,W) = f(W,V) = \sum_{ijk} W_{ij}\left(\hat{\Sigma}_{ij} - (V\hat{\Sigma} V^{\top})_{i,j}\right)$, where the $\hat{\Sigma}$ are empirical covariance matrices obtained from samples distributed according to $\mathcal{N}(0,\Sigma)$.
For the experiments, the matrix $\Sigma$ is created as $\Sigma = U U^T$, where the entries of $U \in \mathbb{R}^{d \times d}$ are distributed i.i.d. standard Gaussian.\\

\textbf{Replication and Findings:}
As described in the original paper, this project considers the algorithms OGDA, SGA, ConOpt, and CGD for deterministic case $\hat{\Sigma} = \Sigma$, corresponding to the limit of large sample size, with $\gamma = 1.0$, $\epsilon = 10^{-6}$. Let dimensions of $\Sigma$, $d \in \{20, 40, 60\}$ and let the stepsizes range over $\eta \in \{0.005, 0.025, 0.1, 0.4\}$. \\
Authors suggest to evaluate the algorithms according to the trade-off between the number of forward evaluations and the corresponding reduction of the residual $\|W+W^{\top}\|_{\operatorname{FRO}}/2+ \|UU^{\top} - V V^{\top}\|_{\operatorname{FRO}}$, starting with a random initial guess (the same for all algorithms) obtained as $W_{1} = \delta W$, $V_{1} = U + \delta V$, where the entries of $\delta W, \delta V$ are i.i.d uniformly distributed in $[-0.5,0.5]$.
Authors count the number of "forward passes" per outer iteration as follows.
\begin{itemize}
    \item OGDA: 2
    \item SGA: 4
    \item ConOpt: $6$
    \item CGD: 4 + 2 $*$ number of CG iterations 
\end{itemize}

\begin{figure}
    \centering
    \includegraphics[scale=0.215]{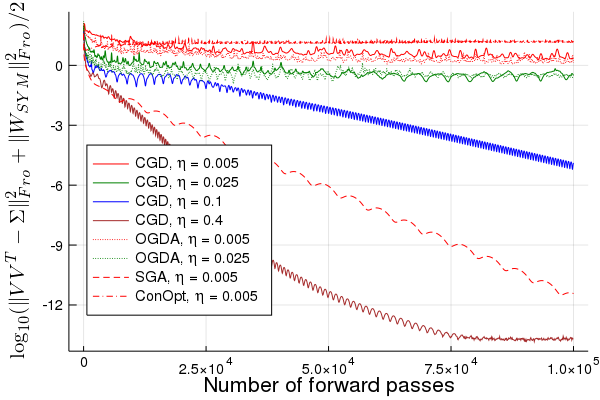}
    \includegraphics[scale=0.215]{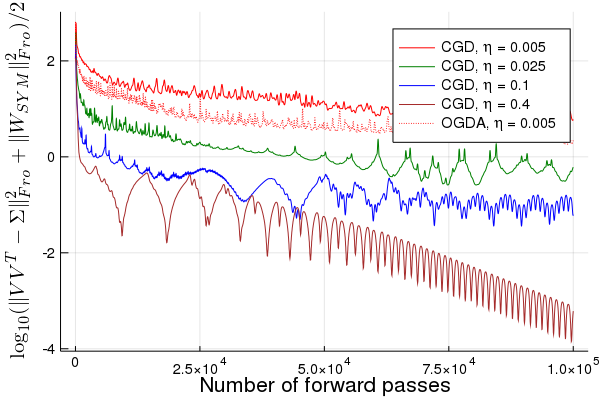}
    \includegraphics[scale=0.215]{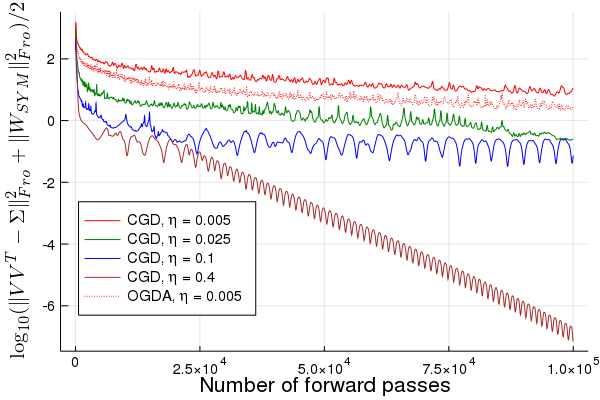}
    \caption{Original Paper result:}

    \label{fig:matrixDet}

    \includegraphics[scale=0.215]{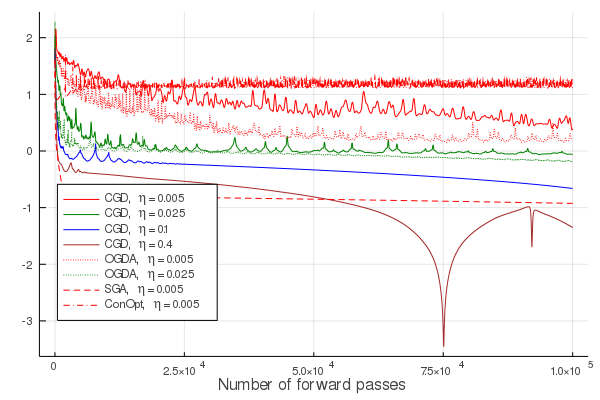}
    \includegraphics[scale=0.215]{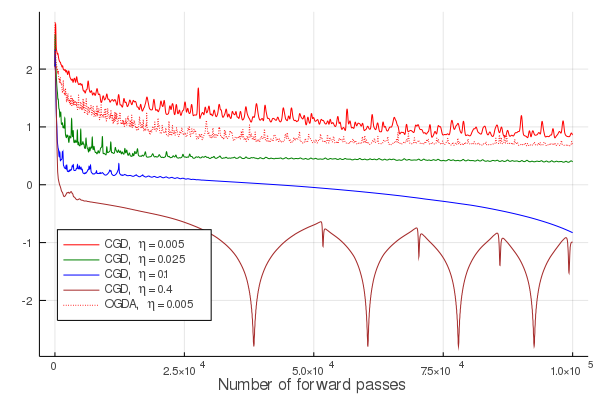}
    \includegraphics[scale=0.215]{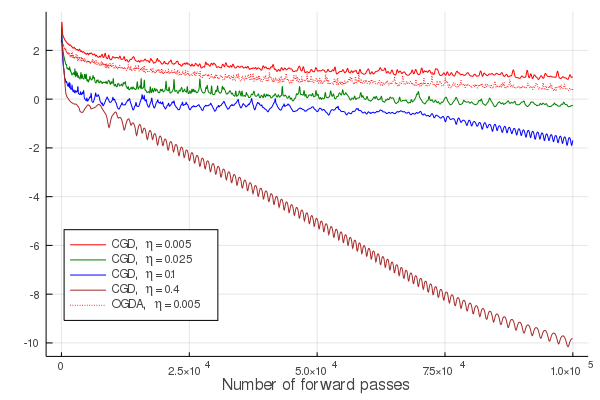}
    \caption{Replication result: Plot of the decay of the residual after a given number of model evaluations, for increasing problem sizes and $\eta \in \{0.005, 0.025, 0.1, 0.4\}$.}

    \label{fig:mymatrixDet}
\end{figure}

The results are summarized in Figure~\ref{fig:matrixDet} and Figure~\ref{fig:mymatrixDet}. 
We see consistently that for the same stepsize, CGD has convergence rate comparable to that of OGDA.
However, as we increase the stepsize the other methods start diverging, thus allowing CGD to achieve significantly better convergence rates by using larger stepsizes. 
For larger dimensions ($d\in \{40, 60\}$) OGDA, SGA, and ConOpt become even more unstable such that OGDA with the smallest stepsize is the only other method that still converges, although at a much slower rate than CGD with larger stepsizes.

\textbf{Conclusion :}
On studying the tradeoff between the number of evaluations of the forward model and the residual, this project supports that larger stepsize, the convergence rate of CGD is better than the other methods and for comparable stepsize, the convergence rate of CGD is similar to the other methods.

\subsection{Experiment 3: Fitting a bimodal distribution}
To further study the robustness of the CGD, authors propose to use a simple GAN to fit a Gaussian mixture model with two modes, in two dimensions.

\begin{figure}
    \includegraphics[scale=0.333]{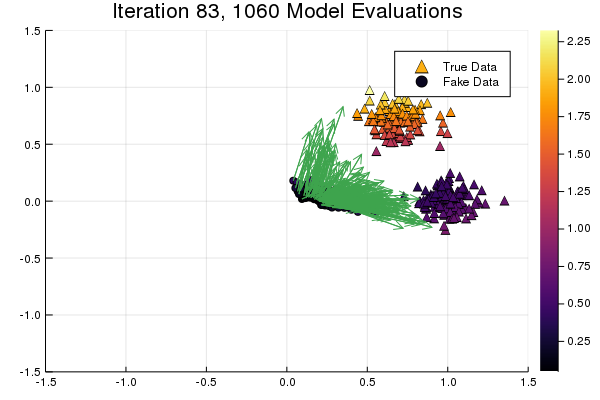}
    \includegraphics[scale=0.333]{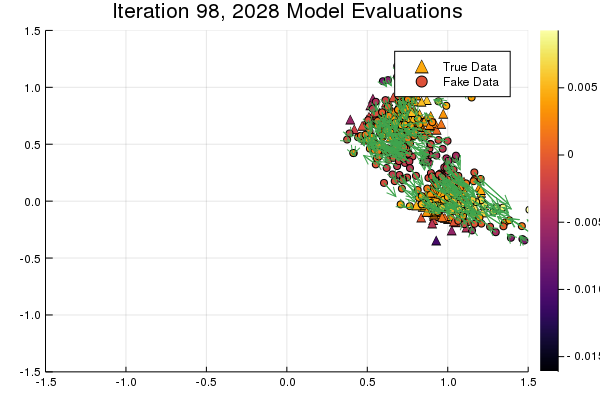}
    \caption{Original Paper result:}

    \label{fig:two_mode_gan}
    

    \includegraphics[scale=0.333]{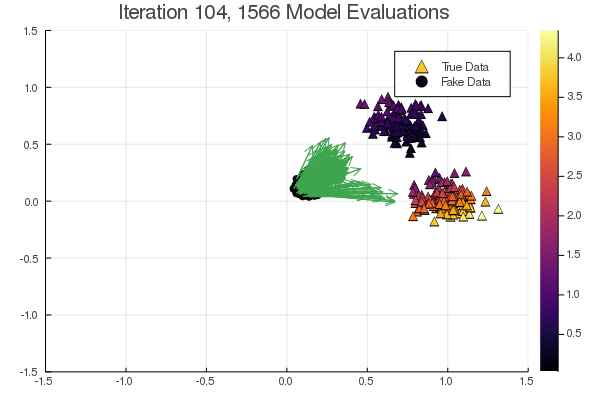}
    \includegraphics[scale=0.333]{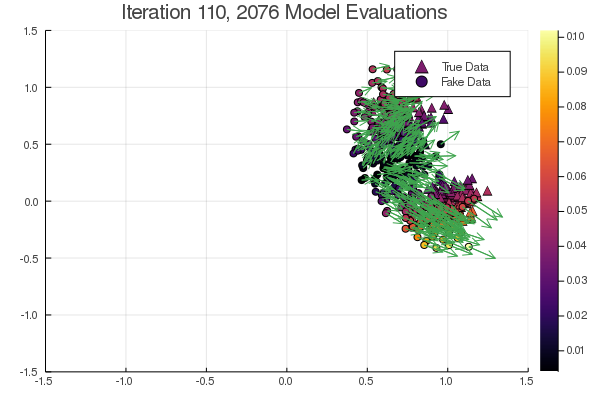}
    \caption{Replication result: Under CGD, the mass eventually distributes evenly among the two modes.}

    \label{fig:my_two_mode_gan}

\end{figure}

\textbf{Replication and Findings}
We use a GAN to fit a Gaussian mixture of two Gaussian random variables with means $\mu_{1} = (0,1)^{\top}$ and $\mu_{2} = (2^{-1/2}, 2^{-1/2})^{\top}$, and standard deviation $\sigma = 0.1$\\
Generator and discriminator are given by dense neural nets with four hidden layers of $128$ units each that are initialized as orthonormal matrices, and ReLU as nonlinearities after each hidden layer. 
The generator uses 512-variate standard Gaussian noise as input, and both networks use a linear projection as their final layer. 
At each step, the discriminator is shown 256 real, and 256 fake examples.
We interpret the output of the discriminator as a logit and use sigmoidal crossentropy as a loss function.The results are summarized in Figure~\ref{fig:two_mode_gan} and Figure~\ref{fig:my_two_mode_gan}. \\

We apply SGA, ConOpt ($\gamma = 1.0$), OGDA, and CGD for stepsize $\eta \in \{0.4, 0.1, 0.025, 0.005\}$ together with RMSProp ($\rho = 0.9)$.
In each case, CGD produces an reasonable approximation of the input distribution without any mode collapse.
In contrast, all other methods diverge after some initial cycling behaviour.

\textbf{Conclusion :}
On all methods, the generator and discriminator are initially chasing each other across the strategy space, producing the typical cycling pattern. 
When using SGA, ConOpt, or OGDA, however, eventually the algorithm diverges with the generator either mapping all the mass far away from the mode, or collapsing the generating map to become zero. 
Therefore, we also tried decreasing the stepsize to $0.001$, which however did not prevent the divergence.
For CGD, after some initial cycles the generator starts splitting the mass and distributes is roughly evenly among the two modes.

\textbf{\em This project supports authors' claim that CGD is significantly more robust than existing methods for competitive optimization.}

\section{Conclusion and Acknowledgement}
This project mostly successfully replicates the work of original paper. It supports the properties of introduced \emph{Competitive Gradient Descent} method for Competitive optimization through the proposed experiments. The three experiments test different aspects of the algorithms.
Experiment 1 is based on removal of cyclic behaviour and importance of different definition terms. It also tests convergence/divergence of the algorithms on different zero-sum games.
Experiment 2 evaluates and compares computational complexity of the CGDA w.r.t other algorithms.
Experiment 3 studies the robustness of CGDA by fitting bimodal distribution using GAN.

This project acknowledges the code of the original paper written in Julia used for replication. The original paper describes the experiments very well and their code is also very understandable. It also acknowledges the support of Code Ocean for providing compute resources and prof. Debasish Ghose, Department of Aerospace Engineering, Indian Institute of Science.

\end{document}